\pdfoutput=1

\documentclass[11pt, twocolumn]{article}

\usepackage{acl}
\usepackage{lipsum}
\usepackage{cuted}
\usepackage{graphicx}
\usepackage{tcolorbox}
\tcbuselibrary{breakable}
\tcbuselibrary{skins}
\usepackage{hyperref}
\usepackage{times}
\usepackage{latexsym}
\usepackage{multirow}

\usepackage[T1]{fontenc}

\usepackage[utf8]{inputenc}

\usepackage{microtype}
\usepackage{booktabs}
\usepackage{amssymb}

\title{Learning to Check: Enhancing Self-Correction Capabilities in Large Language Models for Reasoning Tasks}
\usepackage{amsmath}
\usepackage{graphicx}
\usepackage{amsfonts}
\author{Che Zhang\textsuperscript{*} \textsuperscript{$\spadesuit$} \textsuperscript{$\vartriangle$}, Zhenyang Xiao\textsuperscript{*} \textsuperscript{$\spadesuit$}, Chengcheng Han \textsuperscript{$\diamondsuit$} , Yixin Lian\textsuperscript{$\heartsuit$}, Yuejian Fang \textsuperscript{$\spadesuit$} \textsuperscript{$\dagger$}\\
  \textsuperscript{$\spadesuit$} School of Software \& Microelectronics, Peking University \\
  \textsuperscript{$\diamondsuit$}School of Data Science and Engineering, East China Normal University\\
  \textsuperscript{$\heartsuit$}Xiaobing.AI\\
  \texttt{\{mmt, kjn\}@stu.pku.edu.cn, fangyj@ss.pku.edu.cn} \\
  \texttt{chengchenghan@stu.ecnu.edu.cn} \\
  \texttt{lianyixin@xiaobing.ai} \\
  }
\begin{document}
\newcommand{\up}[1]{\textsuperscript{\textuparrow #1}}
\newcommand{\down}[1]{\textsuperscript{\textdownarrow #1}}
\maketitle
\newcommand\blfootnote[1]{%
\begingroup
\renewcommand\thefootnote{}\footnote{#1}%
\addtocounter{footnote}{-1}%
\endgroup
}
\blfootnote{\textsuperscript{*}Equal contribution.}
\blfootnote{\textsuperscript{$\vartriangle$} Work done during an internship at Xiaobing.AI.}
\blfootnote{\textsuperscript{$\dagger$}Corresponding author.}

\begin{abstract}

Self-correction has achieved impressive results in enhancing the style and security of the generated output from large language models (LLMs). However, recent studies suggest that self-correction might be limited or even counterproductive in reasoning tasks due to LLMs' difficulties in identifying logical mistakes. 

In this paper, we aim to enhance the self-checking capabilities of LLMs by constructing training data for checking tasks. Specifically, we apply the Chain of Thought (CoT) methodology to self-checking tasks, utilizing fine-grained step-level analyses and explanations to assess the correctness of reasoning paths. We propose a specialized checking format called \textbf{``Step CoT Check''}.
Following this format, we construct a checking-correction dataset that includes detailed step-by-step analysis and checking. Then we fine-tune LLMs to enhance their error detection and correction abilities.

Our experiments demonstrate that fine-tuning with the ``Step CoT Check'' format significantly improves the self-checking and self-correction abilities of LLMs across multiple benchmarks. This approach outperforms other formats, especially in locating the incorrect position, with greater benefits observed in larger models.

For reproducibility, all the datasets and code are provided in \href{https://github.com/bammt/Learn-to-check}{https://github.com/bammt/Learn-to-check}.
\end{abstract}

\section{Introduction}

The colossal number of parameters and extensive training data of Large Language Models~(LLMs)
have led to remarkable performances in reasoning tasks~\cite{yao_react_2023, wei_chain--thought_2023}, 
notably in mathematical problem-solving~\cite{luo_wizardmath_2023, wang_mathcoder_2023, yu_metamath_2023, yuan_scaling_2023}.
Despite these advancements, 
there remains considerable potential to further enhance the mathematical reasoning capabilities of LLMs.

To further improve the mathematical reasoning capabilities of LLMs, some studies~\cite{madaan_self-refine_2023, weng_large_2023, huang_large_2023, saunders_self-critiquing_2022, welleck_generating_2022, miao_selfcheck_2023, chen_iterative_2023} have attempted to prompt LLMs to identify and self-correct errors in their reasoning paths, thereby achieving higher accuracy. In general, self-correction involves two distinct processes: \textbf{self-checking feedback gathering} and \textbf{feedback-driven reasoning}. The former requires LLMs to review their previous output and attempt to identify errors within it, while the latter allows the model to modify and optimize its previous outputs based on the received feedback.

However, recent research~\cite{huang_large_2023, stechly_gpt-4_2023, valmeekam_can_2023} indicates that without external feedback, 
such as ground truth labels for determining the stopping point~\cite{shinn_reflexion_2023, kim_language_2023},  LLMs cannot improve their reasoning abilities through \textbf{intrinsic self-correction}. This difficulty arises because LLMs often fail to accurately identify errors in reasoning paths and provide precise feedback.~\cite{stechly_gpt-4_2023, an_learning_2023,madaan_self-refine_2023}. This is counterintuitive because, for humans, verifying whether an answer is correct is often easier than directly solving the problem. The former is an NP-hard problem, while the latter is a P problem.  Therefore, we hypothesize that by constructing specific task training, LLMs can enhance their ability to directly check for errors.

Given these challenges, our work aims to address the limitations by constructing a \textbf{checking-correction dataset} and  then fine-tuning LLMs using this dateset.
For generating the dataset, we first collect reasoning paths that contain errors and then we propose a specialized prompt format for checking called \textbf{``Step CoT Check''}. This format involves evaluating the correctness of each step by sequentially checking different types of errors and providing detailed explanations. If an error is detected, the model concludes the check for that step and provides a summary. Following this format, we generated data to check erroneous reasoning paths, and subsequently used these checks to create a correction dataset. Then we fine-tune our models on a mixed dataset of CoT data and checking-correction data.

In our setup, the entire reasoning process includes a stage of direct reasoning and two stages of self-correction: (1) direct reasoning, (2) self-checking feedback gathering and (3) feedback-driven reasoning. Our training data encompasses three different tasks, each corresponding to one of these stages. And in the stage 2, we compared the performance of the Step CoT Check format with two other prompt formats: All Direct Check and Step Direct Check.

Our contributions are threefold:
\begin{enumerate}
    \item We propose using training data specifically designed to check reasoning paths and demonstrate their effectiveness in enhancing the self-checking capabilities. This approach enables LLMs to achieve self-correction independently of external data.
    
    \item We introduce a Chain-of-Thought (CoT) method that incorporates checking tasks. Specifically, we propose the ``Step CoT Check'' format for checking mathematical reasoning tasks, which categorizes error types and elucidates the reasoning process prior to judgment, thereby improving interpretability. This approach markedly surpasses traditional direct answer verification techniques. 

    \item We have released a checking-correction dataset to improve the performance of LLMs on reasoning tasks. To our knowledge, this dataset is the first to use fine-grained, step-by-step explanations to annotate errors in LLMs' outputs.
\end{enumerate}

\section{Related Work}

Chain-of-Thought~(CoT) reasoning, 
initially pioneered by~\citet{wei_chain--thought_2023}, 
has significantly advanced the reasoning capabilities of large language models (LLMs). 
This approach, which enhances LLMs by incorporating detailed reasoning steps into few-shot examples,
has been foundational in improving performance across various reasoning tasks.
Subsequent research has expanded upon this, introducing methodologies like Self-Consistency~\cite{wang2022self}, 
various prompting techniques~\cite{zhou2022least,drozdov2022compositional},
Self-Training~\cite{huang2022largecanselfimprove}, 
Verifier~\cite{li2022advance}, 
Tree of Thought~\cite{yao2023tree}
and
DialCoT~\cite{han2023dialcot}.
The effectiveness of CoT has been demonstrated in domains including logical, commonsense, and mathematical reasoning. 
Despite the success, 
these models are not without their flaws. 
They are still prone to reasoning confusion, generating hallucinations, and computational errors. 
A prevailing strategy to rectify these undesired behaviors of LLMs is learning from feedback.
Extensive research~\cite{huang2022largecanselfimprove,madaan_self-refine_2023,gero2023self,jiang2023selfevolve} has been undertaken to improve LLMs through the paradigm of learning from feedback. 
However,~\citet{huang_large_2023} found that self-improvement is not effective when oracle labels are not available.
This may be attributed to the limited capability of LLMs to identify errors. 
Previous works have primarily focused on self-correction, overlooking the enhancement of the models' error identification capabilities. 
This paper addresses the gap by meticulously constructing training data that includes steps for detecting and correcting errors, 
thereby strengthening the model's capability to identify mistakes.

\section{Methodology}
In this section, we present our methodology, which comprises two components: \textbf{checking-correction data generation} and \textbf{fine-tuning LLMs}. The former component involves collecting erroneous reasoning paths, generating data that includes a detailed checking process and the revised answers. The latter involves fine-tuning the model on a combined dataset that contains three tasks: direct reasoning, checking the correctness of the reasoning paths and pointing out the error locations, and re-reasoning to obtain the correct answers.
\subsection{Checking-Correction Data Generation}
\begin{figure}[htbp]
    \centering
    \includegraphics[width=1.0\linewidth]{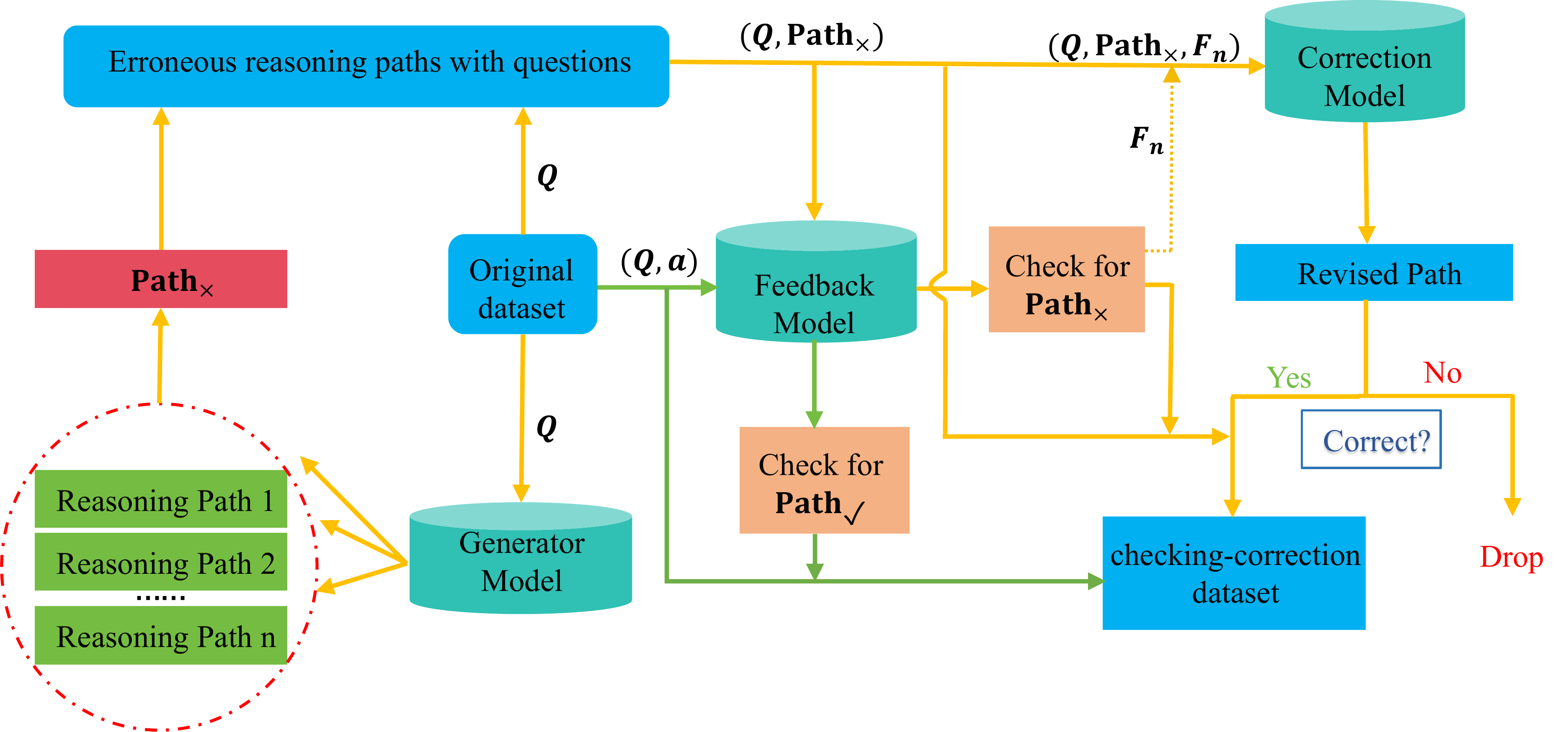}
    \caption{The process of checking-correction dataset generation. }
    \label{fig_2}
\end{figure}

The process of generating the checking-correction dataset is detailed in Figure \ref{fig_2}.

\subsubsection{Collecting Reasoning Paths with Errors}
We use a generator model $\mathcal{M}_g$ to produce reasoning path $P ( Ans(P) \neq a$, and $a$ is the correct answer.) for original question $Q$, then we get pairs $x=(Q, P)$. Like ~\citet{an_learning_2023}, we use a newline character as a delimiter to decompose $P$ into fine-grained reasoning steps $s_1, s_2, \ldots, s_K$, and append “Step N:” before each step.

To simplify the notation, we use $\text{\textbf{Path}}_{\times}$ to represent the reasoning path with logical errors. For comparison, we use $\text{\textbf{Path}}_{\checkmark}$ to represent the reasoning paths that lead to a correct answer.

\subsubsection{Error Type Definition}
To construct a prompt format suitable for LLMs to check the answers to mathematical problems, we categorize and analyze the types of errors in a certain step of the generation $\text{\textbf{Path}}_{\times}$ mentioned above. In mathematical reasoning tasks, the types of errors are as follows:
\begin{enumerate}
    \item \textbf{Goal Errors}: The model outputs the process of calculating a quantity that was already mentioned in the known conditions or a quantity that could not be computed based on the existing known conditions. Alternatively, it calculates an intermediate quantity that is irrelevant to the final problem's solution, ultimately leading to a deviation in the reasoning path.
    \item \textbf{Computational Expression Errors}: The model experiences a hallucination or misunderstands the given conditions of the problem, leading to an incorrect process for calculating an intermediate quantity. This often manifests as the outputted computational expression not matching the given conditions.
    \item \textbf{Calculation Errors}: The model correctly identifies the intermediate quantity that needs to be calculated and outputs the correct computational expression, but the calculation result is incorrect.
\end{enumerate}

\begin{figure*}[htbp]
    \centering
    \includegraphics[width=0.85\textwidth]{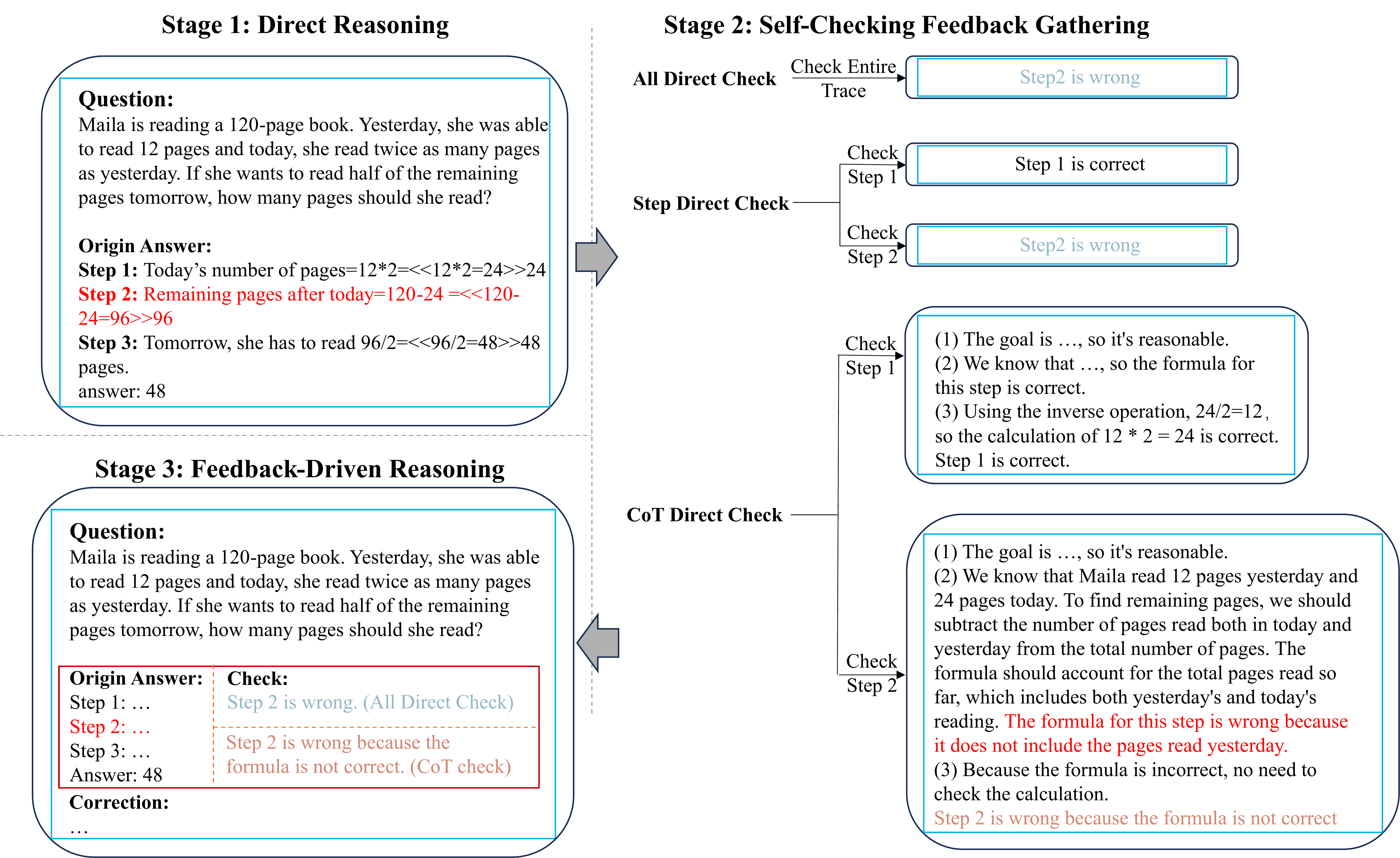}
    \caption{The reasoning process combined with self-correction. The Process involves three stages. \textbf{Stage 1} involves direct reasoning for a given question. \textbf{Stage 2} examines the direct reasoning path and provides feedback at the end. If Stage 2 check identifies errors, Stage 3 is then initiated. \textbf{Stage 3} requires the model to generate a new reasoning path and attempt to correct the erroneous answer given the question, original answer, and feedback.}
    \label{fig_1}
    \vspace{-10pt} %
\end{figure*}

\subsubsection{Checking Data Generation}
For a reasoning path $P$ involving $K$ steps, we prompt a feedback model $\mathcal{M}_f$ to generate checking data for each step $s_k, k \in [1, K]$ one by one. For each step, $\mathcal{M}_f$ check sequentially according to the three types of errors mentioned above and generate a detailed explanation process.
\begin{enumerate}
    \item Reasoning Goal Appropriateness: Assesses whether the computational goal of $s_k$ is conducive to solving the question. If the step's objective is unreasonable or redundant, it is deemed incorrect.
    \item Computational Expression Correctness: Identifies the known conditions related to $s_k$'s computational goal and evaluates if the computation aligns with the problem statement and whether the derived expression achieves the computational goal correctly.
    \item Verification through Inverse Operations: The correctness of the computation is checked by performing inverse operations on the derived expression.
\end{enumerate}
If the above error types are detected, the model will generate a brief summary of the error message; otherwise, it will summary that the step is correct. Since the three types of errors are hierarchical and subsequent errors may depend on the initial error, it only makes sense to check for a particular type of error if the previous types have not occurred. For the same reason, we only generate the checking data of steps up to the first wrong step.\footnote{In practical experiments, redundant steps are also considered correct as long as they do not contain logical errors. Only steps that contain logical errors are used as stopping criteria.} If step $s_n, n \in [1, K]$ is determined to be incorrect after these checks, a concise summary indicating the error will be generated and denoted as $F_n$. We refer to this method of step-by-step checking of arithmetic reasoning paths with detailed explanations as ``\textbf{Step CoT Check}''. 

Before generating data on a large scale, we use GPT-4-1106-preview~\cite{openai_gpt-4_2023} as $\mathcal{M}_f$ and manually assess the quality of 50 checking data samples of $\text{\textbf{Path}}_{\times}$ to ensure data quality. Among these, 43 entries are \textbf{perfect}(accurately identify the erroneous step and provide precise analysis), while 5 entries are \textbf{good} (pinpoint the incorrect step but offer analysis that lacks sufficient accuracy). 2 entries are \textbf{poor} (fail to identify the location of the error). Based on this human evaluation, we suppose the overall quality of check data is sufficient for fine-tuning.

To ensure that the model can effectively learn and distinguish between correct and incorrect reasoning processes and reduce bias for the further fine-tuning stage, in addition to $\text{\textbf{Path}}_{\times}$, we also filter some original answers from the CoT dataset as $\text{\textbf{Path}}_{\checkmark}$ and generate their checking data. \footnote{The specific prompts and examples of generated data can be found in Appendix~\ref{sec:appendix}.}
\subsubsection{Correction Data Generation}

We task a correction model $\mathcal{M}_c$ to generate correction data. Specifically, given $(Q, P, F_n)$, where $P \in \text{\textbf{Path}}_{\times}$, the model is required to generate a revised reasoning path $P_r$. We use regular expressions to compare $P_r$ with the original answer $a$ to question $Q$. If $P_r$'s final answer is correct, we then add the checking data and correction data to the Checking-correction dataset; otherwise, we drop the data sample. For the $\text{\textbf{Path}}_{\checkmark}$'s checking data, we also add it to the Checking-correction dataset.

\subsection{Fine-Tuning LLMs}
After generating the checking-correction data, we fine-tune the LLMs to evaluate whether these models can self correct. To this end, we mainly compare the performance of two fine-tuning settings:
\begin{itemize}
    \item \textbf{Fine-Tuning on CoT Data} We fine-tune the model on GSM8K training data alone. To facilitate comparison with our method, we only use the original train set and don't generate additional reasoning paths for training. We use the model fine-tuned with the original CoT data as a baseline and compare it with our self-correction method.
    \item \textbf{Fine-Tuning on CoT Data + Checking-Correction data} In addition to CoT data, we also used the Checking-Correction dataset for training. To eliminate the effect of data size, we employed a replacement method, substituting one piece of data from the original CoT dataset with the corresponding Checking-Correction data for the same question $Q$. 
\end{itemize}

The entire reasoning process is illustrated in Figure~\ref{fig_1}. And in the stage of \textbf{Self-Checking Feedback Gathering}, we employed three different prompt formats and compared the performance of the fine-tuned models:
\begin{itemize}
    \item \textbf{All Direct Check:} Input the original question and the original reasoning path into the model $\mathcal{M}$, instructing $\mathcal{M}$ to check the entire reasoning path. The model should directly output the first step that contains a logical error.

    \item \textbf{Step Direct Check:}  In each inference round, input the original question and the original reasoning path into the model $\mathcal{M}$, ask $\mathcal{M}$ to check one step $s_k, k \in [1, K]$. The complete path $P$ is input to provide $\mathcal{M}$ with a holistic view. The model stop the process as soon as it finds $s_n$ containing an error, then we get the error feedback $F_n$. 

    \item \textbf{Step CoT Check:} The input is the same with ``Step Direct Check''. But before outputs a checking conclusion for each step, this method requires $\mathcal{M}$ to analyze step $s_k$ for correctness from three dimensions as mentioned before. 
\end{itemize}

The input-output format of CoT data and correction data used for fine-tuning are showed in the Appendix~\ref{Appendix:self-checking}. For All Direct Check and Step Direct Check formats, we use regular expressions and rules to generate their responses.

\section{Experiments}

\subsection{Experiment Step}

\paragraph{CoT Dataset}
Our research focuses on multi-step arithmetic reasoning problems and employs the GSM8K dataset~\cite{cobbe_training_2021}. GSM8K contains high quality linguistically diverse grade school
math word problems with 7,473 examples in the training set and 1,319 test cases.

\paragraph{Generate the Checking-Correction Dataset}
We use LLaMA-7b and LLaMA-13b~\cite{touvron_llama_2023} as the generator model $\mathcal{M}_g$ to generate $\text{\textbf{Path}}_{\times}$, which is fine-tuned for one epoch on the GSM8K training set to be able to learn the CoT format. Subsequently, we increase the temperature of $\mathcal{M}_g$ and generate five different responses for each question in the training set. For each question, we filter one reasoning path that contains an incorrect answer.

We use GPT-4-1106-preview~\cite{openai_gpt-4_2023} as $\mathcal{M}_f$ to generate accurate $\text{\textbf{Path}}_{\times}$'s checking data. However, generating $\text{\textbf{Path}}_{\checkmark}$'s checking data is relatively simple, requiring only the "reading comprehension" of the problem conditions and each step in the reasoning path. To save costs, we use GPT-3.5-turbo~\cite{gpt35turbo} as $\mathcal{M}_f$ to generate $\text{\textbf{Path}}_{\checkmark}$'s checking data and as $\mathcal{M}_c$ to generate correction data. 

As a result, There are approximately 4,500 instances of $\text{\textbf{Path}}_{\times}$'s checking and correction data, and 2,000 instances of $\text{\textbf{Path}}_{\checkmark}$'s checking data in the checking-correction dataset.

\paragraph{Fine-Tuning and Evaluation}
We conduct experiment on LLaMA-2~\cite{touvron_llama2_2023}
models with 7B and 13B parameters, trained on 4 Nvidia A100 80GB GPUs. The models undergo full parameter fine-tuning, employing a learning rate of $8 \times 10^{-6}$, over three epochs, with a cumulative batch size of 32. For models trained on mixed data, we used mixed data and replaced the original 4,000 GSM8K entries with checking-correction data. After extensive experimentation, we determined the optimal ratio of $\text{\textbf{Path}}_{\checkmark}$ to $\text{\textbf{Path}}_{\times}$ for each checking format. For the ``All Direct Check'' and ``Step Direct Check'' methods, we adopted a ratio of 3000:1000, while for the ``Step CoT Check'' method, we chose a ratio of 3700:300. (more information about the data ratio is showed in \ref{data proportion}).

The effectiveness of the checkpoints is evaluate on the GSM8K benchmark and additional three datasets: MultiArith~\cite{roy_solving_2015}, ASDiv~\cite{miao_diverse_2020}, and SVAMP~\cite{patel_are_2021}. Each dataset contains elementary mathematical problems with distinct lexical distributions. Our evaluation strategy validates the robustness of our method across these varied data distributions.  We use the full 1,319-instance GSM8K set, and randomly sample subsets from MultiArith (400), ASDiv (500), and SVAMP (500). 

Checkpoint evaluations are conducted using the vLLM library~\cite{kwon_efficient_2023}. For direct reasoning path generation, we set the temperature parameter to zero. The check and correction mechanism employs a temperature of 0.5, complemented by a top-p value of 0.85 and a top-k setting of 30. For all reasoning tasks, we apply a repetition penalty of 1.2~\cite{artiges_replicator_2019}, which effectively mitigates the generation of duplicate content.

\subsection{Main Results and Analysis}

\begin{table*}[h]
\centering
\scalebox{0.9}{
\begin{tabular}{cccccccccc}
\hline
\multirow{2}*{Method}            & \multirow{2}*{Model}      & \multicolumn{2}{c}{GSM8K}                    & \multicolumn{2}{c}{ASDiv}                    & \multicolumn{2}{c}{MultiArith}               & \multicolumn{2}{c}{SVAMP}                    \\
~  &~    &  D-R &w/S-C                    &  D-R &w/S-C                    &  D-R  &w/S-C               &  D-R &w/S-C                    \\ \hline
CoT Fine-Tuning           & \multirow{4}*{LLaMA-2-7B}  & 32.6  &    /            & 26.8   &  /             & 73.5    &  /            & 35.0   &   /             \\
All Direct Check  & ~  & 27.2 &32.5 \up{5.3}       & 25.2  & \textbf{31.6}  \up{6.4}         & 71.3 &76.8 \up{5.5}         & 32.0 & 35.8 \up{3.8}        \\
Step Direct Check & ~  & 25.1 &32.4 \up{7.3}          & 26.2 &29.2 \up{3.0}         & 72.0 &79.0 \up{7.0}          & 30.0 &35.6 \up{5.6}          \\
Step CoT Check    & ~ & 27.9 & \textbf{34.8} \up{6.9} & 26.0 &31.0 \up{5.0} &72.8 & \textbf{80.3} \up{7.5} & 31.6 &\textbf{36.4} \up{4.8} \\ \hline
CoT Fine-Tuning               & \multirow{4}*{LLaMA-2-13B} & 45.6    &  /            & 45.2    &   /           & 85.3    &   /           & 47.2    &    /         \\
All Direct Check  & ~ & 43.5 &48.1 \up{4.6}          & 41.4 &45.8 \up{4.4}          & 87.0 & 89.5 \up{2.5}          & 47.0 & 50.8 \up{3.8}          \\
Step Direct Check & ~ & 43.7 & 48.8 \up{5.1}         & 41.2 &45.4 \up{4.2}          & 82.5 & 86.5 \up{4.0}          & 45.0 & 49.2 \up{4.2}          \\
Step CoT Check    & ~ & 45.3 & \textbf{51.2} \up{5.9} & 44.2 & \textbf{49.0} \up{4.8} & 88.3 & \textbf{92.0} \up{3.7} & 48.2 & \textbf{53.0} \up{4.8}\\ \hline
\end{tabular}
}
\caption{\label{correct_rate}
Correct rates (\%) on the test set combined with the self-correction mechanism. For each example in the table, the number on the left represents the correct rate of direct reasoning(D-R), and the number on the right represents the correct rate after self-correction(S-C).
}
\end{table*}

\label{sec:main_res}

\paragraph{Performance Of Reasoning Combined Self-Correction}

\begin{figure}[]
    \centering
    \includegraphics[width=\columnwidth]{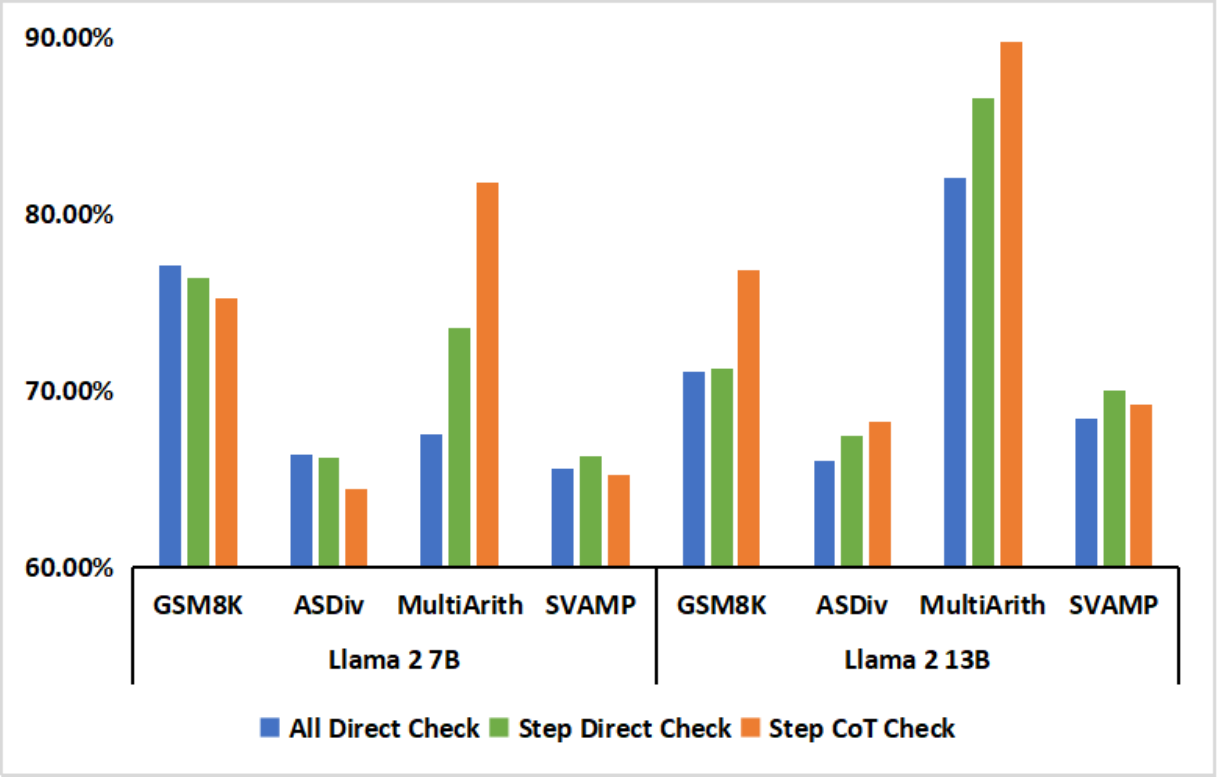}
    \caption{Accuracy($\%$) of checking the correctness of reasoning paths obtained by itself.}
    \label{fig:ck_acc}
\end{figure}
Table~\ref{correct_rate} and Figure~\ref{fig:ck_acc} illustrate the performance of models trained using our three prompt formats on the test sets. When a portion of the CoT data is replaced with checking-correction instances, there is a decrease in the models' direct reasoning accuracy. This decrease is more pronounced in the LLaMA-2-7B model, while it is much smaller in the LLaMA-2-13B model. This suggests that models with larger parameters may be able to obtain extra information from the Checking-Correction data, thereby somewhat improving CoT performance \cite{an_learning_2023}.

For each dataset, all models show a significant increase in accuracy following self-correction. And in almost all datasets, models trained on mixed data outperform those fine-tuned solely on CoT data. Notably, compared to the COT baseline, the improvement in the LLaMA-2-13B model is more significant than in the LLaMA-2-7B model, with the former showing an average increase of 5.5\% and the latter 3.8\%. This discrepancy is primarily due to a larger reduction in direct reasoning performance in the LLaMA-2-7B model following data substitution. In the LLaMA-2-13B model, the "Step CoT Check" method significantly outperforms other checking methodologies, indicating that in models with larger parameters, categorizing error types and explicitly displaying the checking process can provide more precise feedback.

\paragraph{Comparison of Checking Abilities.}
\begin{table*}
\centering
\begin{tabular}{cccccc}
\hline
Method            & Model              & $\text{\textbf{Path}}_{\checkmark}$    & $\text{\textbf{Path}}_{\times}$    & $Average$            & $Step$              \\ \hline
\multirow{4}*{Few-Shot}            & GPT-3.5-turbo*      & 93.0 & 47.0 & 70.0          & / \\
~            & GPT-4-1106-preview* & 85.0 & 93.0 & 89.0          & / \\ 
~            & LLaMA-2-7B& 74.0 & 35.0 & 54.5         & / \\
~            & LLaMA-2-13B& 69.0 & 49.0 & 59.0           & / \\ \hline
All Direct Check  & \multirow{3}*{LLaMA-2-7B} & \textbf{84.0} & 63.0 & \textbf{73.5}          & 28.0  \\
Step Direct Check & ~ & 80.0 & 65.0 & 72.5          & 22.0           \\
Step CoT Check  & ~ & 63.0 & \textbf{77.0} & 70.0 & \textbf{30.0}           \\ \hline
All Direct Check  & \multirow{3}*{LLaMA-2-13B} & 61.0 & \textbf{83.0} & 72.0          & 36.0           \\
Step Direct Check & ~ & 76.0 & 79.0 & 77.5          & 32.0           \\
Step CoT Check  & ~ & \textbf{80.0} & \textbf{83.0} & \textbf{81.5} & \textbf{48.0}  \\ \hline
\end{tabular}
\caption{\label{check-result}
The accuracy (\%) of checking sampled 200 reasoning paths. $Step$  denotes the rate of successfully identifying the first wrong steps in 100 $\text{\textbf{Path}}_{\times}$.
}
\vspace{-10pt} %
\end{table*}

To fairly evaluate the model's capability of checking CoT traces and finding errors, we sampled 200 reasoning traces from GSM8K and manually annotated the first incorrect step. This sample included 100 $\text{\textbf{Path}}_{\checkmark}$ and 100 $\text{\textbf{Path}}_{\times}$. In addition to our fine-tuned models, we also conducted experiments on the models without fine-tuning, as well as on two closed-source LLMs: GPT-3.5-turbo and GPT-4-1106-preview. When using the few-shot method to prompt these models, the original model may fail to follow complex prompts and output incorrect steps. Therefore, we required the model to directly evaluate the entire reasoning path to determine its correctness. When generating arithmetic reasoning paths, we find that the probability of obtaining the correct answer after a logical error occurs is extremely low. Therefore, when checking the reasoning path step by step, if any step is deemed incorrect, we mark the entire reasoning path as incorrect. Only when all steps are correct do we mark the entire reasoning path as correct.

As shown in Table \ref{check-result}, the original LLaMA models performed poorly in checking, just above random (50\%). Although the performance on GPT-3.5 and GPT-4 seems quite good (with an accuracy of 70\% for the former and 89\% for the latter), it is worth noting that in direct reasoning scenarios, the accuracies of GPT-3.5 and GPT-4 are 75.9\% and 95\%~\cite{tyen_llms_2024}, respectively. If the model's ability to judge correctness is comparable to its ability to directly output the correct answer (assuming the model's accuracy in judging the correctness of a reasoning path on questions beyond its capability range is 50\%), then the accuracy of checking should be 89.95\% and 97.5\%. It is evident that in LLMs that have not undergone specialized training, the ability to verify a reasoning path is significantly lower than the ability to directly output the correct answer to a question. We believe that it is precisely this "mismatch" between direct reasoning and checking abilities that leads to ineffective intrinsic self-correction.

In all of our fine-tuned models, the average accuracy exceeds 70\%, which is significantly better than the untrained base models. This indicates that through training, the models' self-checking capabilities have been enhanced. Notably, the LLaMA-2-13B model performs exceptionally well in the ``Step CoT Check'' method, achieving an 81.5\% accuracy in path checking and demonstrating a higher success rate in identifying the initial incorrect step, significantly outperforming other methods. However, on the LLaMA-2-7B model, although the probability of identifying the initial incorrect step is relatively high, its accuracy in checking the path does not show a significant advantage compared to the other two formats, suggesting that similar to CoT reasoning\cite{wei_chain--thought_2023}, CoT for checking is also more effective in models with larger parameters.

Despite the LLaMA-2-7B model's lower probability of accurately identifying specific erroneous steps, self-correction still leads to significant improvements in the previous experiments. This is because, although the model may not precisely pinpoint the error, it can still recognize incorrect answers, thus providing numerous incorrect paths with opportunities for regeneration and subsequent correction.

\subsection{Ablation Study}

To examine the propensity of LLMs to commit specific errors and to validate the effectiveness of our "Step CoT Check" three-step checking method, we conducted an ablation study focusing on the ``Step CoT Check'' implemented on the LLaMA-2-13B model. The findings are catalogued in Table \ref{ablation study}. Among the 541 error paths correctly detected, 20.1\% of the errors were successfully corrected to be accurate. Calculation errors were the most common, accounting for approximately 61.1\% of the total errors. These errors usually stem from the LLMs' inability to correctly understand the relationships in the problem or the erroneous use of non-existent conditions, leading to the incorrect construction of mathematical expressions. These computational missteps proved challenging to rectify, with a mere 18\% being corrected during the third step, signaling an avenue for future enhancement.

Goal errors, conversely, represent a category of errors more amenable to correction, with a rectification rate of 24\%. Regarding computation, LLMs were found to be more susceptible to miscalculations when the equation involved multiple numerical entities or integrated unknowns.
\begin{table}
\centering
\scalebox{0.72}{
\begin{tabular}{ccccc}
\hline
                        & Goal    & Formula & Calculate   & Total           \\ \hline
Number of wrong cases   & 131     & 329     & 81          & 541             \\
Successfully correction & 31      & 61      & 17          & 109             \\
Performance gains       & 1.74\%  & 3.11\%  & 0.99\%      & 5.84\% \\ \hline
\end{tabular}
}
\caption{\label{ablation study}
The results of ablation study. In the table, Performance gains is the proportion of ``wrong to correct'' minus the number of ``correct to wrong'' cases.
}
\end{table}
\subsection{Further Analysis on Data Proportion}
\label{data proportion}
The experiments in this section use the Step Check prompt, and fine tune on LLama-2-13B.
\begin{table*}[h]
\centering
\begin{tabular}{ccccccc}
\hline
 \multirow{2}*{$\text{\textbf{Path}}_{\times}:\text{\textbf{Path}}_{\checkmark}$}    &  \multirow{2}*{$\text{\textbf{Step}}_{\times}:\text{\textbf{Step}}_{\checkmark}$}    &  \multirow{2}*{Ratio} & \multicolumn{3}{c}{Check Accuracy($\%$)}  & \multirow{2}*{Correct Rate($\%$)}  \\
 ~ & ~ & ~ & $\text{\textbf{Path}}_{\checkmark}$                & $\text{\textbf{Path}}_{\times}$                & $Average$                 & ~ \\ \hline
2000:2000 & 2000:9260 & 0.22 & \textbf{93.6} & 36.2          & 60.8          & 45.3          \\
3000:1000 & 3000:7225 & 0.42 & 88.0          & 47.0          & 64.5          & 47.5          \\
3500:500  & 3500:6117 & 0.57 & 84.5          & 61.8          & 72.2          & 50.3          \\
3700:300  & 3700:5681 & 0.65 & 78.9          & 75.0          & \textbf{76.8} & \textbf{51.2} \\
4000:0    & 4000:4990 & 0.80 & 46.1          & \textbf{88.1} & 69.2          & 48.6           \\ \hline
\end{tabular}
\caption{\label{tab:porprotion}
Results of the data scaling experiment. The table displays how the number of correct and incorrect reasoning paths in the training data have an effect on the accuracy of the check and the correct rate of the reasoning combined with self-correction.
}
\vspace{-10pt} %
\end{table*}
\paragraph{The check ability has a relationship with the data ratio.} To discern the influence of the proportion of $\text{\textbf{Path}}_{\checkmark}$ confirmations to $\text{\textbf{Path}}_{\times}$ corrections within a pool of 4,000 training data replacements, we adjusted their ratios and documented the outcomes in Table \ref{tab:porprotion}. On average, a $\text{\textbf{Path}}_{\checkmark}$ comprised 3.33 correct steps ($\text{\textbf{Step}}_{\checkmark}$) checking, whereas a $\text{\textbf{Path}}_{\times}$ included an average of 1.24 correct steps ($\text{\textbf{Step}}_{\checkmark}$) checking and 1.0 wrong steps ($\text{\textbf{Step}}_{\times}$) checking. When the ratio of $\text{\textbf{Step}}_{\checkmark}$ to $\text{\textbf{Step}}_{\times}$ is too high, the model may tend to judge the steps as correct, thereby creating bias. Conversely, an excess of incorrect-checking data may lead the model to judge answers as incorrect more frequently, thereby increasing the instances of "correct becoming wrong," and consequently reducing performance.

From the table, it can be observed that maintaining a ratio of approximately 0.65 between $\text{\textbf{Step}}_{\times}$ and $\text{\textbf{Step}}_{\checkmark}$ in the training set allows the model to achieve a high level of accuracy in distinguishing between correct and incorrect steps. This equilibrium facilitated the highest performance enhancement through a single iteration of the self-correction process.

\begin{figure}[]
    \centering
    \includegraphics[width=\columnwidth]{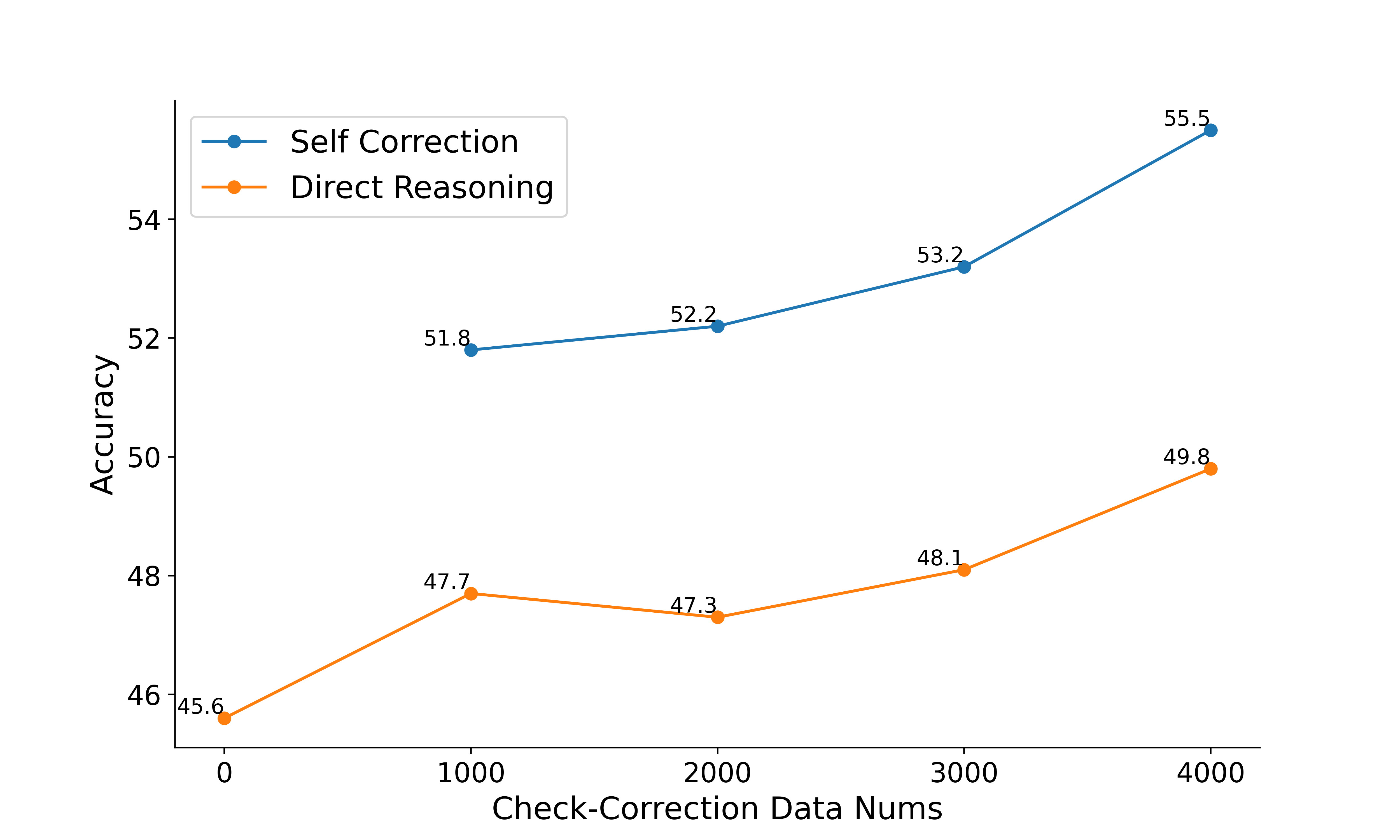}
    \caption{Accuracy~($\%$) of direct reasoning and reasoning combined with self-correction on GSM8K as checking-corection data types increase.}
    \label{fig:proportion}
\end{figure}

\paragraph{More checking-corection data can continuously improve the reasoning capability of LLMs.} Unlike previous experiments, we use the full GSM8K training dataset and add our own checking-correction data. We aim to see how increasing checking-correction data affects the model's reasoning, keeping the original reasoning data the same. We adhere to the optimal ratio of $\text{\textbf{Path}}_{\checkmark}$ and $\text{\textbf{Path}}_{\times}$ ascertained from preceding experiments. Fig. \ref{fig:proportion} shows that both direct and self-corrected reasoning improve as we increase the checking-correction data.

Similar to the insights from LEMA~\cite{an_learning_2023}, We found that LLMs can also enhance their direct CoT reasoning abilities by training on our checking-correction dataset. Additionally, an increased corpus of checking-correction data markedly bolsters the model's self-correction proficiency. Combining reasoning with self-correction improves faster than direct reasoning alone. These results endorse the inclusion of error-annotated correction data within the training regimen of an inference model, positing that an expanded dataset catalyzes improved inferential processes within the model. Consequently, the model's self-corrective capacity is significantly enhanced, culminating in substantial performance advancements through the mechanism of self-correction.
\section{Conclusion}

In this paper, 
we focused on improving the self-correction capabilities of LLMs by enhancing their self-checking abilities.
By conducting an in-depth analysis of error patterns in mathematical reasoning, 
we developed the ``Step CoT Check'' prompt and constructed a checking-correction dataset. This dataset, combined with the original CoT data, has been instrumental in training models that exhibit improved self-checking and self-correction,
reducing dependence on external feedback or ground truth labels for error identification and correction. 
Notably, models fine-tuned with our ``Step CoT Check'' prompt demonstrated superior performance over those trained with alternative prompts.
This approach not only enhances the precision of feedback but also significantly increases correctness rates, 
showing the effectiveness of our method in refining LLMs for complex reasoning tasks.

\clearpage
\section*{Limitations}
Our paper has some limitations, which we leave for future work:
\paragraph{The computational cost of the step level Check} Compared to direct reasoning and checking the whole path at once, although step-by-step check leads to better performance, it requires more inference rounds. On the GSM8K dataset, inference using step-by-step checking combined with correction process requires 4.8 model calls compared to only one for direct reasoning and 2.6 for checking the whole path at once. In practical use, there is a trade-off between performance and resource consumption.
\paragraph{Lack of applicability to broader reasoning tasks.} We analyze the types of mistakes in mathematical reasoning and propose a set of hints for checking the reasoning process. Our approach focuses on checking and correcting errors in mathematical reasoning problems, and it remains to be investigated in future work how to extend this approach to a wider range of reasoning tasks.
\paragraph{Lack of experimentation on larger data sizes and scale language models.} 
Due to constraints on computational resources and funding, we do not conduct experiments on larger scale language models(>20B), or generate additional CoT and correction data like other works. Thus the performance on larger LLMs and sufficient amount of training data remains undetermined. We will further explore the performance of our method on larger data sizes and scale language models
in future research.
\section*{Ethics Statement}
The proposed method has no obvious potential
risks. All the scientific artifacts used/created are
properly cited/licensed, and the usage is consistent with their intended use. Also, we open up
our codes and hyper-parameters to facilitate future
reproduction without repeated energy cost.
\clearpage

\bibliography{custom}
\bibliographystyle{acl_natbib}
\newpage
\appendix
\onecolumn

\section{Prompt for Checking-corection Data Generation}
\label{sec:appendix}
\begin{tcolorbox}[enhanced jigsaw, 
  width=\textwidth, 
  breakable, 
  boxsep=2pt, 
  arc=0pt, 
  outer arc=0pt,
  boxrule=1pt, 
  colback=white, 
  colframe={rgb:red,4;green,2;blue,3}, 
  coltext=black, 
  title=Example 1: Prompt for Checking Data Generation
]

\textbf{System:}\\
Below you will be given a [Question] and the [original answer] of it which has mistakes. You start from the first step, think carefully to check step by step. For each step, you should strictly follow three processes to check: (1) Find out the goal of this step in the [original answer], and you check if the goal is reasonable, i.e., it’s not included in the known conditions, can be calculated from previous information, and it can help to solve the final question. Notice you don’t need to check the calculation process here. (2) Analyze the what we know now and how to calculate the goal, then determine if the formula in [original answer] is correct. Notice you should analyze before you check! (3) Use the inverse operation to check if the calculation is correct (e.g. 17-9=8 to check 8+9=17 and 25-9-8=8 to check 8+8+9=25 ). And when you find the first wrong step, stop check and print ``Stop check! ''. \\

\textbf{Human:}\\
\text{[Question]:} Maila is reading a 120-page book. Yesterday, she was able to read 12 pages and today, she read twice as many pages as yesterday. If she wants to read half of the remaining pages tomorrow, how many pages should she read?\\
\text{[original answer]:} \\
Step 1: Today's number of pages = 12 x 2= <<12*2=24>>24\\
Step 2: Remaining pages after today = 120 - 24 = <<120-24=96>>96\\
Step 3: Tomorrow, she has to read 96 / 2 = <<96/2=48>>48 pages.\\
answer: 48\\
**reference answer**: \\
Maila read 12 x 2 = <<12*2=24>>24 pages today.\\
So she was able to read a total of 12 + 24 = <<12+24=36>>36 pages since yesterday.\\
There are 120 - 36 = <<120-36=84>>84 pages left to be read.\\
Since she wants to read half of the remaining pages tomorrow, then she should read 84/2 = <<84/2=42>>42 pages.\\
\#\#\#\# 42

\tcblower 

\textbf{Assistant:} \\
Step 1: \\
(1) The goal is calculate the number of pages Maila read today. Because it contributes to solving the final problem, so it’s reasonable.\\
(2) We already know that she read twice as many pages today as she did yesterday. To get today’s pages, we should multiply yesterday's reading by 2 . The formula for this step is correct. \\
(3) Using the inverse operation, 24/2=12, so the calculation of 12 * 2 = 24 is correct. \\
So step 1 is correct.\\
Step 2: \\
(1) The goal is calculate the remaining pages after today. Because it contributes to solving the final problem, so it’s reasonable. \\
(2) We already know that Maila read 12 pages yesterday and 24 pages today. To find remaining pages, we should subtract the number of pages read both in today and yesterday from the total number of pages. The formula should account for the total pages read so far, which includes both yesterday's and today's reading. The formula for this step is wrong because it does not include the pages read yesterday. \\
(3) Because the formula is incorrect, no need to check the calculation. \\
So step 2 is wrong because the formula is not correct.\\
Stop check!

\end{tcolorbox}

\clearpage

\begin{tcolorbox}[enhanced jigsaw, 
  width=\textwidth, 
  breakable, 
  boxsep=2pt, 
  arc=0pt, 
  outer arc=0pt,
  boxrule=1pt, 
  colback=white, 
  colframe={rgb:red,4;green,2;blue,3}, 
  coltext=black, 
  title=Example 2: Prompt for Correction Data Generation
]

\textbf{System:}\\
You are an AI assistant who can check and correct the answers very well. Please correct the [original answer] of the [Question] to get the correct answers. You should correct the error pointed out by the [check] in the [original answer]. Please keep your reasoning process close to the [original answer] if possible.\\
\textbf{Human:}\\
\text{[Question]:} Julie is reading a 120-page book. Yesterday, she was able to read 12 pages and today, she read twice as many pages as yesterday. If she wants to read half of the remaining pages tomorrow, how many pages should she read? \\
\text{[original\_answer]:} \\
step 1: Today's number of pages = 12 x 2= <<12*2=24>>24\\
step 2: Remaining pages after today = 120 - 24 = <<120-24=96>>96\\
step 3: Tomorrow, she has to read 96 / 2 = <<96/2=48>>48 pages.\\
answer: 48\\
\text{[check]:} \\
Step 2 is wrong because the formula is not correct.\\

\tcblower 

\textbf{Assistant:} \\
The original answer is wrong because it incorrectly calculated the remaining pages after today in Step 2. The correct reasoning process is as follows:\\
Step 1: Today's number of pages = 12 x 2 = 24\\
step 2: Remaining pages after today is 120 - 12 - 24 = <<120-12-24=84>>84\\
step 3: Tomorrow, she has to read 84 / 2 = <<96/2=48>>42 pages.\\
answer: 42

\end{tcolorbox}

\section{Prompt for Reasoning Combined with Self-correction Mechanisms}
\label{Appendix:self-checking}
\begin{tcolorbox}[enhanced jigsaw, 
  width=\textwidth, 
  breakable, 
  boxsep=2pt, 
  arc=0pt, 
  outer arc=0pt,
  boxrule=1pt, 
  colback=white, 
  colframe={rgb:red,4;green,2;blue,3}, 
  coltext=black, 
  title=Example 3: Direct Reasoning
]

\textbf{System:}\\
You are an AI assistant for solving math problems that can think step by step calculate the results accurately.\\
\textbf{Human:}\\
Natalia sold clips to 48 of her friends in April, and then she sold half as many clips in May. How many clips did Natalia sell altogether in April and May? \\
\tcblower 

\textbf{Assistant:} \\
Step 1: Natalia sold 48/2 = <<48/2=24>>24 clips in May.\\
Step 2: Natalia sold 48+24 = <<48+24=72>>72 clips altogether in April and May.\\
answer: 72

\end{tcolorbox}

\begin{tcolorbox}[enhanced jigsaw, 
  width=\textwidth, 
  breakable, 
  boxsep=2pt, 
  arc=0pt, 
  outer arc=0pt,
  boxrule=1pt, 
  colback=white, 
  colframe={rgb:red,4;green,2;blue,3}, 
  coltext=black, 
  title=Example 4: Feedback-Driven Reasoning
]

\textbf{System:}\\
You are an AI assistant who can check and correct the answers very well. Please correct the [original answer] of the [Question] to get the correct answers. You should correct the error step pointed out by the [check] in the [original answer]. Please keep your reasoning process close to the [original answer] if possible.\\
\textbf{Human:}\\
\text{[Question]:} Tina makes \$18.00 an hour.  If she works more than 8 hours per shift, she is eligible for overtime, which is paid by your hourly wage + 1/2 your hourly wage.  If she works 10 hours every day for 5 days, how much money does she make?\\
\text{[original answer]:} \\
Step 1: She works 8 hours a day for \$18 per hour so she makes 8*18 = \$<<8*18=144.00>>144.00 per 8-hour shift\\
Step 2: She works 10 hours a day and anything over 8 hours is eligible for overtime, so she gets 10-8 = <<10-8=2>>2 hours of overtime\\
Step 3: Overtime is calculated as time and a half so and she makes \$18/hour so her overtime pay is 18*.5 = \$<<18*.5=9.00>>9.00\\
Step 4: Her overtime pay is 18+9 = \$<<18+9=27.00>>27.00\\
Step 5: With the overtime pay included it gives her base earnings of 144+27 =\$<<144+27=171.00>>171.00 per five day week\\
answer: 171\\
\text{[check]:} \\
Step 5 is wrong. \textbf{(All Direck Check and Step Direct Check)}\\
Step 5 is wrong because the formula is not correct.\textbf{(Step CoT Check)}
\tcblower 

\textbf{Assistant:} \\
Step 1: Tina works 8 hours a day for \$18 per hour, so she makes 8 * \$18 = \$144 per 8-hour shift.\\
Step 2: Tina works 10 hours a day, and anything over 8 hours is eligible for overtime, so she gets 10 - 8 = 2 hours of overtime.\\
Step 3: Overtime is calculated as time and a half, so she makes \$18 * 1.5 = \$27 per hour for overtime.\\
Step 4: Her overtime pay is \$27 * 2 = \$54.\\
Step 5: With the overtime pay included, her base earnings of \$144 and her overtime pay of \$54, she makes \$144 + \$54 = \$198 per day.\\
Step 6: Tina works 5 days a week, so her total earnings per week are \$198 * 5 = \$990.\\
Answer: \$990
\end{tcolorbox}

\begin{tcolorbox}[enhanced jigsaw, 
  width=\textwidth, 
  breakable, 
  boxsep=2pt, 
  arc=0pt, 
  outer arc=0pt,
  boxrule=1pt, 
  colback=white, 
  colframe={rgb:red,4;green,2;blue,3}, 
  coltext=black, 
  title=Example 5: Feedback-Acquisition(Direct Check)
]

\textbf{System:}\\
You are an AI assistant who can check and correct the answers very well. Please check if the [solution] to the following question is correct. If the answer is not correct, point out the error step.
\textbf{Human:}\\
\text{[Question]:} Weng earns \$12 an hour for babysitting. Yesterday, she just did 50 minutes of babysitting. How much did she earn?\\
\text{[solution]:} \\
Step 1: Weng earns 12/60 = \$<<12/60=0.2>>0.2 per minute.\\
Step 2: Working 50 minutes, she earned .2*50 =\$ <<.2*50=10>>10.\\
answer: 10\\
\tcblower 
\textbf{Assistant:} \\
The answer is all correct.\\
\end{tcolorbox}

\begin{tcolorbox}[enhanced jigsaw, 
  width=\textwidth, 
  breakable, 
  boxsep=2pt, 
  arc=0pt, 
  outer arc=0pt,
  boxrule=1pt, 
  colback=white, 
  colframe={rgb:red,4;green,2;blue,3}, 
  coltext=black, 
  title=Example 6: Feedback-Acquisition(Direct Step)
]

\textbf{System:}\\
You are an AI assistant who can check and correct the answers very well. Please check if the [solution] to the following question is correct. If the answer is not correct, point out the error step.
\textbf{Human:}\\
\text{[Question]:} Weng earns \$12 an hour for babysitting. Yesterday, she just did 50 minutes of babysitting. How much did she earn?\\
\text{[solution]:} \\
Step 1: Weng earns 12/60 = \$<<12/60=0.2>>0.2 per minute.\\
Step 2: Working 50 minutes, she earned .2*50 =\$ <<.2*50=10>>10.\\
answer: 10\\
Please determine if Step 1 is correct.
\tcblower 

\textbf{Assistant:} \\
Step 1 is all correct.
\end{tcolorbox}

\begin{tcolorbox}[enhanced jigsaw, 
  width=\textwidth, 
  breakable, 
  boxsep=2pt, 
  arc=0pt, 
  outer arc=0pt,
  boxrule=1pt, 
  colback=white, 
  colframe={rgb:red,4;green,2;blue,3}, 
  coltext=black, 
  title=Example 7: Feedback-Acquisition(CoT Step)
]

\textbf{System:}\\
You are an AI assistant who can check and correct the answers very well. Based on the given text, please check if the mentioned step of the [solution] is correct.\\
\textbf{Human:}\\
\text{[Question]:} Betty is saving money for a new wallet which costs \$100. Betty has only half of the money she needs. Her parents decided to give her \$15 for that purpose, and her grandparents twice as much as her parents. How much more money does Betty need to buy the wallet?\\
\text{[solution]:} \\
Step 1: In the beginning, Betty has only 100 / 2 = \$<<100/2=50>>50.\\
Step 2: Betty's grandparents gave her 15 * 2 = \$<<15*2=30>>30.\\
Step 3: That leaves Betty with 50 + 30 = \$<<50+30=80>>80 from all sides.\\
Step 4: So now, Betty still lacks 100 - 80 = \$<<100-80=20>>20.\\
answer: 20\\
Please determine if Step 1 is correct.\\
\tcblower 

\textbf{Assistant:} \\
Step 1:\\
The goal is to determine how much money Betty initially has. This is reasonable because it's necessary to calculate the total amount she will have after receiving additional money from her parents and grandparents.\\
From the question, we know that Betty has only half of the money she needs for the wallet, which costs \$100. To find out how much she has, we divide the total cost by 2. The formula for this step is correct.\\
Using the inverse operation, 50 * 2 = 100, so the calculation of 100 / 2 = 50 is correct.\\
Step 1 is correct.
\end{tcolorbox}
\end{document}